\documentclass[conference]{IEEEtran}
\IEEEoverridecommandlockouts
\usepackage{cite}
\usepackage{amsmath,amssymb,amsfonts}
\usepackage{algorithmic}
\usepackage{graphicx}
\usepackage{textcomp}
\usepackage{xcolor}
\usepackage{cite}
\usepackage{epstopdf}
\usepackage{enumerate}
\usepackage{float}
\usepackage{subfig}
\usepackage{caption}
\usepackage{setspace}
\usepackage{url}

\def\BibTeX{{\rm B\kern-.05em{\sc i\kern-.025em b}\kern-.08em
    T\kern-.1667em\lower.7ex\hbox{E}\kern-.125emX}}
\begin{document}

\title{Potential UAV Landing Sites Detection \\through Digital Elevation Models Analysis}

\author{\IEEEauthorblockN{ Efstratios Kakaletsis, Nikos Nikolaidis }
\IEEEauthorblockA{\textit{Department of Informatics} \\\textit{ Artificial Intelligence and Information Analysis Laboratory } \\
\textit{Aristotle University of Thessaloniki}\\
GR-54124 Thessaloniki, GREECE \\
Emails: \{ekakalets, nnik\}@csd.auth.gr}
}

\maketitle

\begin{abstract}
In this paper, a simple technique for Unmanned Aerial Vehicles (UAVs) potential landing site detection   using terrain information through identification of flat areas, is presented. The algorithm utilizes digital elevation models (DEM) that represent the height distribution of an area. Flat areas which constitute appropriate landing zones for UAVs in normal or emergency situations result by thresholding the image gradient magnitude of the digital surface model (DSM). The proposed technique also uses connected components evaluation on the thresholded gradient image in order to discover connected regions of sufficient size  for landing. Moreover, man-made structures and vegetation areas are detected and excluded from the potential landing sites. Quantitative performance evaluation of the proposed landing site detection algorithm in a number of areas  on real world and synthetic datasets, accompanied by a comparison with a state-of-the-art algorithm, proves its efficiency and superiority.  
\end{abstract}

\section{Introduction}
Commercial low cost UAVs (Unmanned Aerial Vehicles), also known as drones,  equipped with several sensors, have found various applications such as in television and filming, search and rescue, surveillance, inspection, mapping, wildlife monitoring, crowd monitoring/management etc \cite{1},\cite{2}. \par
Maps play a crucial role in UAV navigation. Maps can include terrain information that can be used in order to navigate and control the UAV in normal and emergency situations. Such information can come from digital elevation models (DEM) \cite{12} which are produced in two forms. The first is digital terrain models (DTM)\cite{4},\cite{11} that include information regarding the height variations of an area's bare ground without any man-made structures or vegetation. The second type is digital surface models (DSM)\cite{3},\cite{13}. A DSM provides a representation of the elevation values for areas of exposed ground, road surfaces, tree crowns, vegetation and buildings. In other words, DSMs include information for both the ground and the man-made structures or vegetation that lie on it. DSMs can be generated by data coming from various sources such as LiDAR (Light Detection And Ranging) surveying. DTMs are usually generated by post-processing DSMs. DSMs and DTMs often come in raster format i.e. essentially georeferenced images where a pixel's value denotes elevation of the corresponding location. It should be noted here that the terms DEM, DSM, DTM are often used with different definitions than the ones used in this paper. \par
A crucial part of a UAV flight is safe landing. Identifying potential landing areas, which in general should be flat enough, sufficiently large and not occupied by vegetation or buildings by exploiting the aforementioned terrain information, is important both for normal and emergency landing. The literature referring to landing site detection utilizing terrain information, is not very extensive. The authors in \cite{5} achieve landing site detection for fixed-wing UAVs in emergency situations by using the average height and height variance inside quadtree based DEM partitions. Partitions whose height variance is below a limit are selected as landing sites and merged with neighboring partitions if they have similar average heights. Furthermore, the authors use two path planners (the Rapidly-exploring Random Trees (RRT) and the Particle Swarm Optimization (PSO)) - for path planning in limited time from the current UAV position to the closest detected landing site. In \cite{6}, the authors determine suitable landing areas on topographical maps for emergency landing of UAVs by utilizing surface fitting on coarse elevation models using Least Squares Error and slope calculation. Furthermore, in \cite{15}, the authors create a system for efficient and reliable safety assessment of landing zones covered by low vegetation, combining a volumetric occupancy map with a 3D Convolutional Neural Network (CNN). However,  deep learning approaches  require a very large number of training examples which might not be available in certain applications. In contrast, our algorithm requires no training. In \cite{16} the authors propose a system for landing zone selection based on a relatively simple geometric analysis of terrain roughness and slope. Finally, \cite{17} proposes a scheme for the selection or validation of landing zones for unmanned helicopters with terrain assessment incorporating factors such as terrain/vehicle interaction, wind direction and mission constraints.  \par

In this paper, we present a simple but efficient algorithm for UAVs potential landing site detection. The main aim is the identification of (sufficiently) flat and large areas in the topographical maps for UAVs safe landing in normal or emergency situations. The proposed technique utilizes the information in DSM and DTM raster files in order to detect the vegetation, buildings and generally the objects upon the bare ground by simply evaluating the height difference between the DTM and the DSM  model. In addition, flat areas are discovered by evaluating the local terrain slope through the use of an image gradient operator on the DEM file and by thresholding the resulting image in order to keep areas with small slope. Moreover, connected components analysis is used on the resulting binary image in order to find and retain regions whose area is above the minimum potential landing area size. The final result of our algorithm is a list of  sufficiently large map areas with no buildings/vegetation and small terrain slope constituting areas which can be characterized as landing zones. The main advantage of the proposed approach is  that there is no need for complicated training and the respective data. The only prerequisite is the existence of the DSM and DTM height maps, which are publicly available with sufficient resolution for large parts of the globe.

The results of the proposed algorithm are evaluated quantitatively using appropriate metrics (precision and recall) upon both real and synthetic terrain data and manually or semi-automatically derived ground truth data. Moreover, its results are compared to those obtained by the algorithm in \cite{5}, a Digital Elevation Model (DEM) - oriented method, proving the proposed method's superiority. 
 \par
It should be noted here that map-based landing site detection can provide only information regarding potential landing sites, based on terrain geometry. Such landing sites can be precomputed and be available as annotations on the map. Whether these potential landing sites can actually be used for UAV landing at a certain time instance depends on whether the site is free from water, people/crowds or cars at the actual landing time. Such a check can be done either by the pilot through the drone video feed or by applying a person/crowd/car detection algorithm on the drone video.

The remainder of this paper is organized as follows. In Section II, we describe the details of the proposed method. In Section III we present the experiments which have been conducted to measure the algorithm's performance. Finally, conclusions are presented in Section IV.

\section{Potential Landing Site Detection Algorithm}
\begin{figure}[t]
\centering
\includegraphics[width=3.5in]{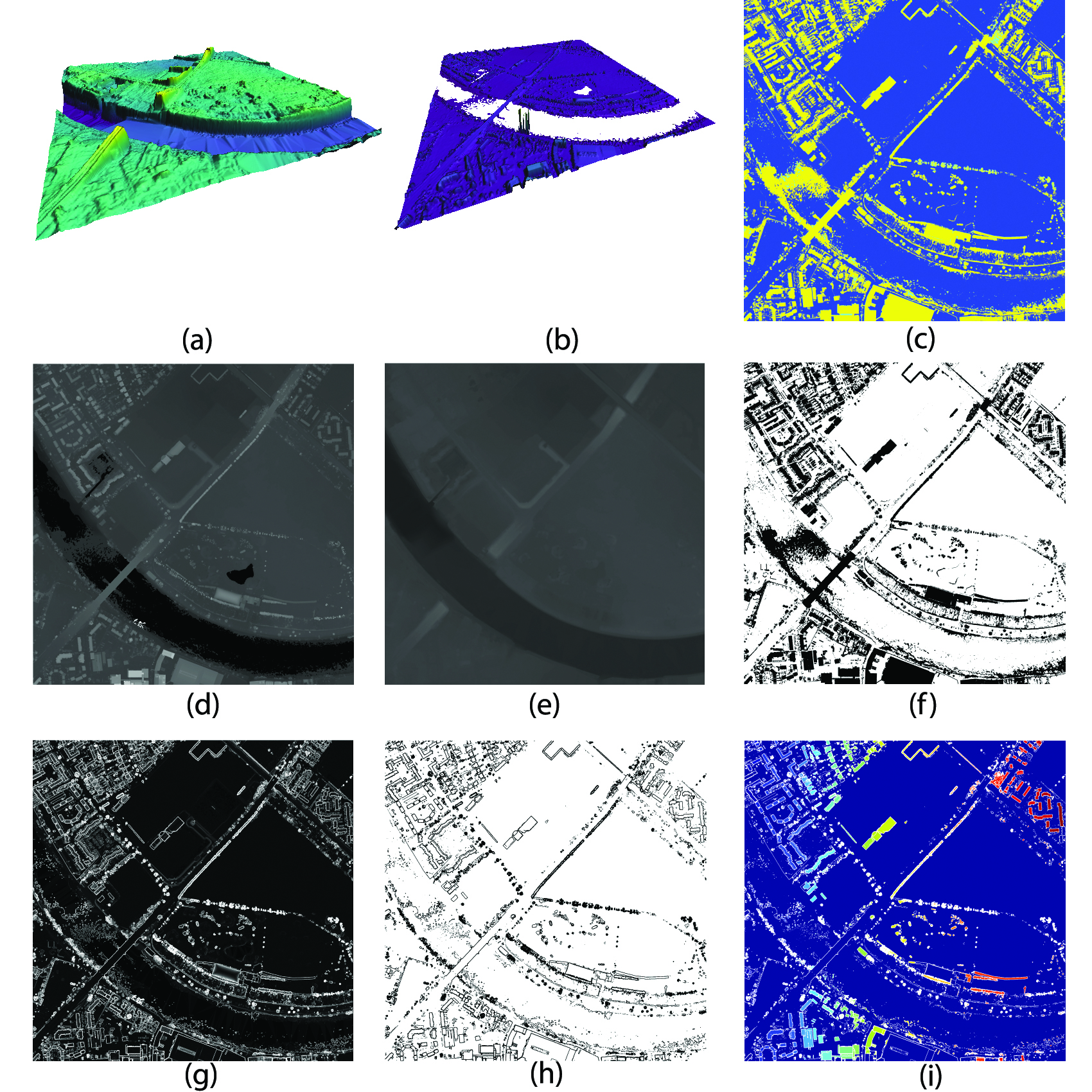}
\centering
\caption{Real world dataset (map1). (a) 3D view of DTM, (b) 3D view of DSM, (c) final map (meaning of colors is explained in the text) (d) DSM, (e) DTM , (f) binary image of buildings/vegetation, in black, (g) Terrain slope determination via Sobel operator, (h) binary image of low slope areas, in white, (i) connected components analysis result.}
\label{fig:vis3}
\end{figure}

 The algorithm's input consists of two digital elevation models, namely the digital surface model (DSM) and the digital terrain model (DTM) of a region in raster format, i.e., as a regular grid of elevation values of a depicted terrain. 
As already mentioned, the DTM (Figure \ref{fig:vis3}-a) depicts just the terrain and no man-made structures or vegetation whereas the DSM (Figure \ref{fig:vis3}-b) depicts the terrain along with buildings and vegetation. It should be noted here that DSM files often contain pixels with no value (no elevation information) which result from sensor inefficiencies during DSM acquisition. As DTM is constructed by post-processing the DSM, these pixels are usually assigned values through some sort of interpolation. In the approach presented below DSM pixels with no values are assigned elevation values from the corresponding pixels of the DTM file. As already mentioned, the algorithm's output is a map that depicts the potential landing zones for the UAV. The algorithm comprises of the five  steps listed below.\par
\begin{figure}[H]
\centering
\includegraphics[width=3.5in]{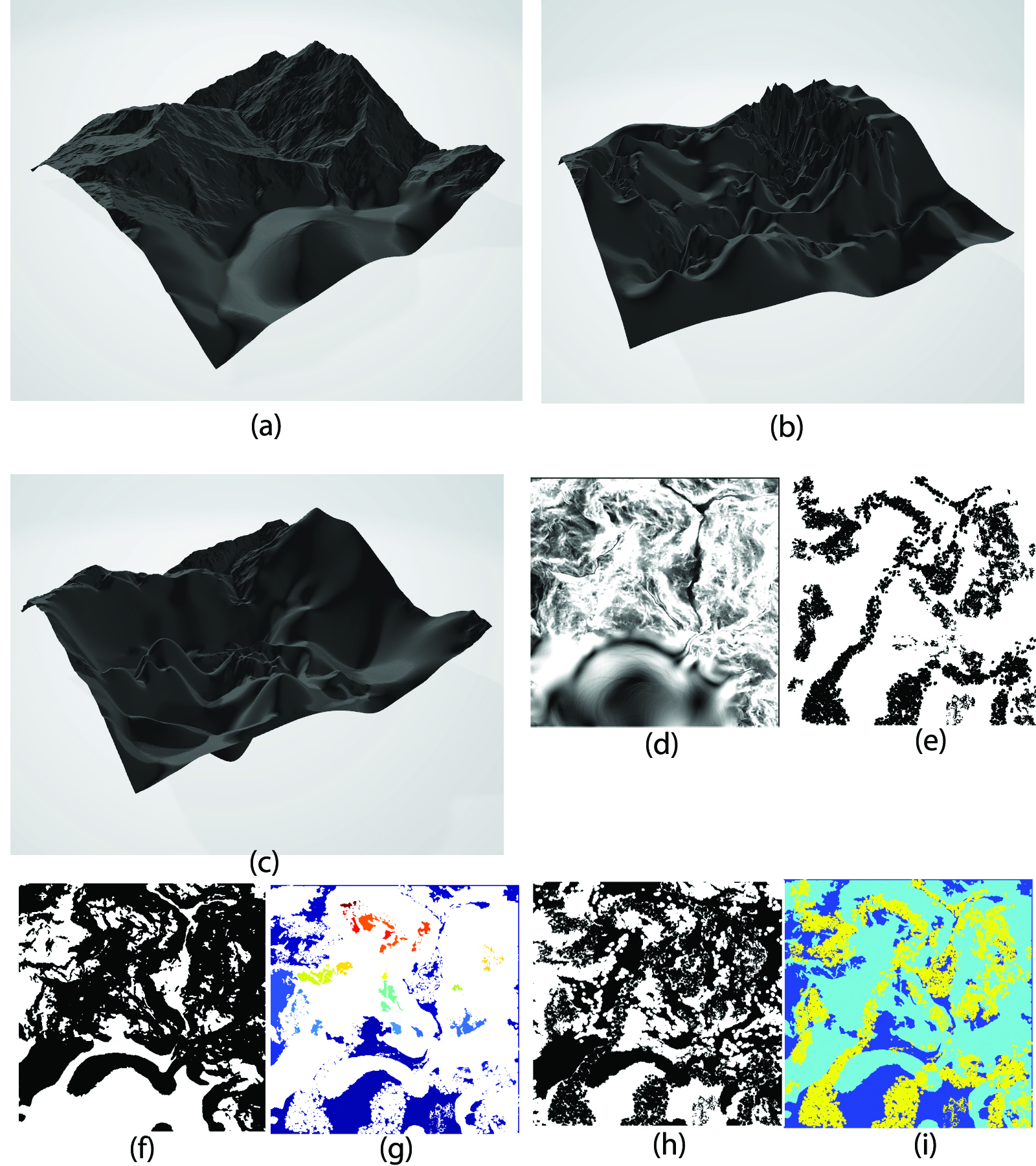}
\centering
\caption{Synthetic Dataset. (a) 3D view of map4 (sub-images (d)-(i) refer to this map), (b) 3D view of map5, (c) 3D view of map6,  (d) Terrain slope determination via Sobel operator, (e) binary image of vegetation, in black, (f) binary image of low slope areas, in white, (g) connected components analysis result, (h) ground truth information, potential landing sites in white (i) final map (meaning of colors is explained in
the text)}
\label{fig:vis4}
\end{figure}

1) Detection of man-made structures and vegetation: By subtracting  DTM from DSM and applying a threshold to the outcome we derive a binary image (Figures \ref{fig:vis3}-f, \ref{fig:vis4}-e) which marks pixels depicting man-made structures and vegetation whose height is above a selected (small) threshold.\par
2) Terrain slope determination (Figures  \ref{fig:vis3}-g,  \ref{fig:vis4}-d): Subsequently, we calculate the local slope of the depicted areas in the DSM. According to Geographic Information System (GIS) literature \cite{14}, slope is the maximum rate of change in value (elevation) from a pixel (cell) to its neighbors. The lower the slope value, the flatter the terrain. As far as the slope calculation is concerned, the rates of change of the surface elevation in the horizontal ($\frac{dz}{dx}$) and vertical ($\frac{dz}{dy}$) directions from the central cell determine the slope. 
Slope, in degrees, is calculated as \cite{14}:

\begin{equation}
slope_{degrees} = \frac{180}{\pi} \arctan \sqrt{\left ( \left [\frac{dz}{dx}  \right ]^{2} +\left [\frac{dz}{dy}  \right ]^{2}  \right ) } 
\end{equation}

\begin{figure}[H]
\centering
\includegraphics[width=0.8in]{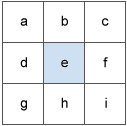}
\centering
\caption{ 8-neighborhood of a DSM.}
\label{fig:neighbor} 
\end{figure}

The values of the center cell and its eight neighbors determine the horizontal and vertical rates of elevation change. For a neighbourhood such as the one depicted in Figure \ref{fig:neighbor} the rates of change in the x and y direction for cell 'e' can be calculated as:
\begin{equation}
\frac{dz}{dx} = \frac{(c + 2f + i) - (a + 2d + g) }{8 * x_{cellsize}}
\end{equation}
\begin{equation}
\frac{dz}{dy} = \frac{(g + 2h + i) - (a + 2b + c)}{8 * y_{cellsize}}
\end{equation}

Essentially, the rates of change in the x and y direction as used in the Geographic Information System (GIS) literature are the horizontal and vertical derivative approximations generated by the well known Sobel operator \cite{9}, scaled by a factor of  $-8*cellsize$.\par  





3) Sobel operator gradient image thresholding (Figure \ref{fig:vis3}-h, \ref{fig:vis4}-f): After extracting the elevation gradient magnitude image, we threshold it in order to classify the DSM pixels in flat  or non-flat areas based on the local slope. Obviously, near flat areas are retained as potential landing areas. The thresholding utilizes a predefined global terrain slope threshold based on slopes which are appropriate for UAVs landing.\par

4) Binary image connected components evaluation (Figure \ref{fig:vis3}-i \ref{fig:vis4}-g): Connected components analysis is applied on the binary image resulting from the previous step. Connected components with sufficiently large number of pixels, i.e. of sufficient area, are retained. 

5) Creation of the final map (Figures \ref{fig:vis3}-c, \ref{fig:vis4}-i): In order to create the final map, we remove from the large, low slope areas found in the previous step those parts that overlap with buildings and vegetation found in step 1. The final map consists of three categories of pixels:
\begin{itemize}
\item Blue pixels: This category of pixels corresponds to the landing zones i.e. the regions in the DSM map which are characterized from small terrain slope and large enough  area for UAV landing.
\item Light blue pixels: This category of pixels corresponds to no landing zones, i.e. the regions in the DSM map with large terrain slope or very few pixels (small area). 
\item Yellow pixels: these pixels also correspond to no landing zones due to buildings and vegetation.
\end{itemize}


\section{Experimental Evaluation}
\subsection{Dataset}

Experimental evaluation of the aforementioned algorithm was conducted in a real world and a synthetic dataset. The first dataset consists of a number of areas  depicted in the DEM data from the publicly available dataset provided by UK’s Environment Agency \cite{10}. This dataset includes digital elevation models from the UK, covering  urban, suburban, rural and bush areas, in spatial resolutions (pixel size per dimension) ranging from 0.25m to 2m. We have selected three areas for our evaluation. The first two  areas (namely map1 depicted along with algorithm results in Figure \ref{fig:vis3} and map2, both of resolution 0.25m) refer to an urban environment with  many structures that prohibit landing, such as buildings and trees. The third examined area (area map3, resolution 2m) is a rural environment with steep downhill  parts.  \par
In addition, the synthetic dataset contains three  3D landscapes (map4, map5, map6) generated  using Unreal Engine 4 (UE4)~\cite{karis2013real}. UE4 is a game engine developed by Epic Games that can achieve high-quality photorealistic graphics  and provides flexible world and asset editors. Using UE4, we created three mostly mountainous landscapes with   different amounts of vegetation and exported the heightmap in raster format as depicted in Figure  \ref{fig:vis4}. The spatial resolution of these data was 1.2m. \par

For the areas in the real world dataset, ground truth (potential landing sites, areas not suitable for landing) was manually constructed 
by the authors through visual inspection of the DEMs and satellite images (the latter were obtained by Google Maps). In the  case of the synthetic dataset, the ground truth for the vegetation was created automatically, since the locations where vegetation was inserted were known. The synthetic nature of the terrain allowed us to calculate  local slope information and use it to mark areas with low slope as being appropriate for landing. Small such areas were excluded.

\subsection{Evaluation metrics and procedure}
The evaluation of the algorithm's performance and its comparison to the method proposed in \cite{5}   was conducted using  precision and recall upon the generated ground truth. In our case, precision refers to the percentage of areas identified by the algorithm as landing sites that are indeed (according to the ground truth) landing sites whereas recall refers to the percentage of the actual landing sites that were identified by the algorithm as such. 

Results for the three areas of the real world and the synthetic  dataset are presented in Tables \ref{my-label} and \ref{my-label2} respectively. The variance threshold for the algorithm in \cite{5} was evaluated by experimentation as the one that gave the best results. Results show that the proposed algorithm can identify potential landing sites with very good precision, especially for the synthetic data. Recall values were also very good. It should be noted  that for the application at hand precision is much more important than recall since landing a UAV in an area that is not suitable for doing so is the error that should be avoided since it might lead to crash landings. \par
 Results also show that the proposed potential landing site detection algorithm outperforms the method proposed in \cite{5} in both the real world and the synthetic dataset. The most gains with respect to \cite{5} were observed in precision figures of the synthetic dataset. The mean computational time (in seconds) needed in order to reach a decision for the DSM/DTM pairs 
in the real world and synthetic datasets, is illustrated in Table \ref{my-time}.
\begin{table}[ht]
\centering
\caption{Computation complexity (execution time)}
\label{my-time}
\begin{tabular}{|l|l|l|}
\hline
  method\textbackslash{}dataset       & \begin{tabular}[c]{@{}l@{}}real world\\ dataset\end{tabular} & \begin{tabular}[c]{@{}l@{}}synthetic \\ dataset\end{tabular} \\ \hline
proposed & 48.5957                                                      & 1.4563                                                       \\ \hline
{[}8{]}  & 6.4702                                                       & 1.3845                                                       \\ \hline
\end{tabular}
\end{table} 
These times refer to a computer with Intel(R) Core(TM) i7-6700HQ CPU @ 2.60Ghz processor and 16GB RAM. It shall be noted  that the algorithm in \cite{5} does not take into account man-made structures and vegetation a fact which contributes in it having inferior performance but smaller execution time. Moreover, the level of granularity of the DTM and DSM obviously affects the execution time: large resolutions increase the execution time of the algorithm. 

\begin{table}[ht]
\centering
\caption{Precision and recall for the three areas of the real world dataset for the proposed algorithm and the one in \cite{5} .}
\label{my-label}
\begin{tabular}{|l|l|l|l|l|}
\hline
metric\textbackslash{}map & map1             & map2             & map3             & method   \\ \hline
precision                 & \textbf{0.7595} & \textbf{0.7446} & \textbf{0.8711} & proposed \\ \hline
                          & 0.7251          & 0.7294      & 0.8678          & \cite{5}  \\ \hline
recall                    & \textbf{0.9786} & \textbf{0.9172} & \textbf{0.4386} & proposed \\ \hline
                          & 0.9548          & 0.8883           & 0.2504          & \cite{5}  \\ \hline
\end{tabular}
\end{table}

\begin{table}[ht]
\centering
\caption{Precision and recall for the three areas of the synthetic dataset for the proposed algorithm and the one in  \cite{5}.}
\label{my-label2}
\begin{tabular}{|l|l|l|l|l|}
\hline
metric\textbackslash{}map & map4            & map5            & map6            & method   \\ \hline
precision                 & \textbf{0.9297} & \textbf{ 0.9027} & \textbf{0.8216} & proposed \\ \hline
                          & 0.6169          & 0.8212         &  0.5784         & {[}8{]}  \\ \hline
recall                    & \textbf{0.7213} & \textbf{0.7578}          & \textbf{0.8498}          & proposed \\ \hline
                          & 0.6856          & 0.7412 & 0.8212   & {[}8{]}  \\ \hline
\end{tabular}
\end{table}

\section{Conclusions}
In this paper, a simple novel method for the identification of potential safe landing sites for Unmanned Aerial Vehicles was introduced. The main aim of the proposed technique is the determination of large flat areas not covered by buildings or vegetation in digital elevation models (DEM). Experiments on real world and synthetic scenes showed that the algorithm can detect such areas with good precision and recall. As far as future work is concerned, this includes the incorporation of learning methods in the determination of landing sites as well as inclusion of information coming from color images, e.g. satellite images from  Google Maps. These additions are expected to increase the algorithm performance and expand its usability to e.g. areas where no elevation data are available. \par 
\section*{Acknowledgments}
 The research leading to these results has received funding from the European Union's Horizon 2020 research and innovation programme  under grant agreement number 731667 (MULTIDRONE). This publication reflects only the authors views. The European Union is not liable for any use that may be made of the information contained therein.\par
The authors would like to thank Mr Charalampos Symeonidis for providing the synthetic dataset generated with Airsim simulator within UE4.\par
This is a preprint of the following work: Efstratios Kakaletsis, Nikos Nikolaidis, "Potential UAV Landing Sites Detection through Digital Elevation Models Analysis", published in Proceedings of the 2019 27th European Signal Processing Conference (EUSIPCO) satellite workshop "Signal Processing Computer vision and Deep Learning for Autonomous Systems".  This EUSIPCO satellite workshop is technically co-sponsored by the IEEE Signal Processing Society (SPS) and endorsed by the IEEE SPS Autonomous Systems Initiative (ASI). 
\bibliographystyle{IEEEtran}
\bibliography{ref}

\end{document}